\documentclass[10pt,twocolumn,letterpaper]{article}

\usepackage{cvpr}
\usepackage{times}
\usepackage{epsfig}
\usepackage{graphicx}
\usepackage{amsmath}
\usepackage{amssymb}
\usepackage{algorithm,algpseudocode}

\usepackage[pagebackref=true,breaklinks=true,letterpaper=true,colorlinks,bookmarks=false]{hyperref}

\cvprfinalcopy 

\def\cvprPaperID{9494} 

\ifcvprfinal\fi
\begin{document}

\title{Radius Adaptive Convolutional Neural Network}

\author{Meisam Rakhshanfar}

\maketitle

\begin{abstract}
   Convolutional neural network (CNN) is widely used in computer vision applications.
   In the networks that deal with images, CNNs are the most time-consuming layer of the networks. 
   Usually, the solution to address the computation cost is to decrease the
   number of trainable parameters.
   This solution usually comes with the cost of dropping the accuracy. 
   Another problem with this technique is that usually the cost of
   memory access is not taken into account which results in insignificant
   speedup gain.
   The number of operations and memory access in a standard convolution layer
   is independent of the input content, which makes it limited for certain
   accelerations.
   We propose a simple modification to a standard convolution to bridge this
   gap.
   We propose an adaptive convolution that adopts different kernel sizes
   (or radii) based on the content.
   The network can learn and select the proper radius based on the input
   content in a soft decision manner. 
   Our proposed radius-adaptive convolutional neural network (RACNN) has a
   similar number of weights to a standard one, yet, results show it can reach
   higher speeds.
   Code has been made available at: \url{https://github.com/meisamrf/racnn}.
   
\end{abstract}

\section{Introduction}

\begin{figure}[t]
\begin{center}
   \includegraphics[width=0.8\linewidth]{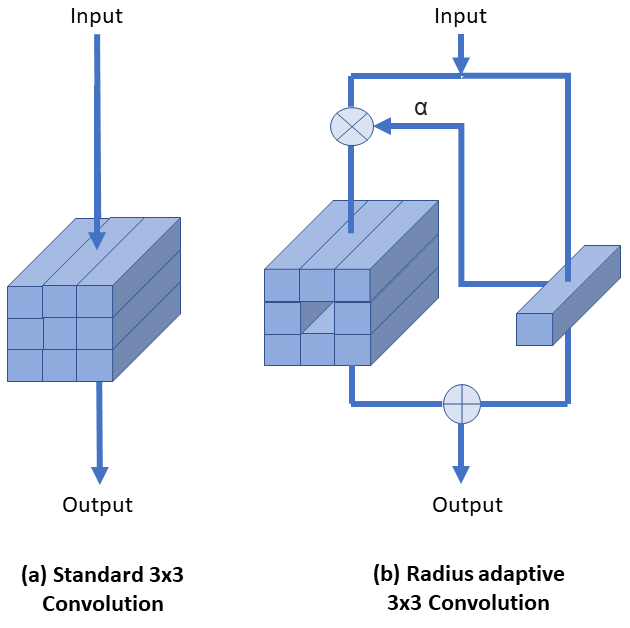}
\end{center}
   \caption{In a radius-adaptive convolution, $\alpha$ defines how
   much of the neighboring pixels are taken into account. $\alpha{=}0$ and
   $\alpha{=}1$ are equivalent to $1{\times}1$ and $3{\times}3$ convolutions.}
\label{fig:main}
\end{figure}

Convolution is a fundamental building block of many deep neural networks. 
In a convolutional layer, a set of per-trained weights extract spatial features
from the input.
For image processing, the input is 3D data in the form of pixels that each
includes certain channels (or depth).
Usually, the total sum of channels for all pixels is a large number. 
Thus, abundant operations and memory accesses are required to perform a
convolution.
Although computers have become more advanced to handle this amount of data, 
computation cost still justifies the need for more speed optimization. 
Many CNN acceleration methods have been introduced so far and except for few, 
most of them are not widely used. This is partially because they are mostly 
application specific and partially because their implementation is not
straight-forward. Some of these approaches require analyzing the weights and network 
after training to compress the network or decompose the weights. For these
reasons, the tendency to use more straightforward techniques like MobileNet
\cite{howard2017mobilenets} is extremely higher.
However, those straightforward techniques usually come with two drawbacks. 
First, their approach to reducing the computation cost is by reducing the
number of parameters (or neurons).
This is not usually desired because it diminishes the accuracy especially for convolution 
layers that the numbers of parameters or weights strongly affect the results. 
Second, they usually do not consider memory access as a slowdown factor. 
However, memory access can be speed-wise costly. As an example, in a separable convolution technique 
used in MobileNet, a 3x3 convolution is divided into two convolutions. Both
convolutions in total have fewer operations compared to a standard one. However,
in each convolution, a read and write from and to the memory is required which
is doubled compared to a standard convolution.
A convolutional layer has a fixed number of operations and memory access. Which
makes its speed content-independent.
One of the advantages that are normally used in traditional speed optimization
is the branching in code,
i.e., if some conditions are met, the program can skip the main processing and
branch to a less-complex processing. As an example, let's consider a text
recognition network that searches for characters in a document. A window of
pixels as an input gets classified as one of the characters or a non-character. There is a high chance that there is no text (or character) in the window. 
Thus, all pixels in the input will have similar values (such as white pixels). 
In such a scenario by checking a simple condition, program can skip the complex 
classification when the condition is met and speedup the process. 
In CNN, implementing such a concept is not as simple as this example
because the building blocks of CNNs are extremely optimized matrix
multiplications. But such scenarios motivated us to seek for a solution.

In this work, we propose a new content-adaptive convolution
that addresses some of the drawbacks of the existing
CNN acceleration ideas. Figure~\ref{fig:main} demonstrates
a high-level scheme of our radius-adaptive convolution for a $3{\times}3$
convolution.
Unlike a standard convolution, the radius of the kernel (or kernel size) can be
adjusted according to the input.
This adjustment is based on a soft decision where the parameter $0{\leq}\alpha{\leq}1$ defines
how much of the neighboring pixels are taken into account. This design is simple to 
implement and the number weights in RACCN 
similar to standard convolution (excluding the weights that select alpha).
When the $\alpha=1$, RACNN acts as a standard $3{\times}3$ convolution and
when $\alpha=0$ it acts as a standard $1{\times}1$ convolution which is the
cause for the speedup.
The remainder of this paper
is as follows: section \ref{sec:related_work} discusses related work; section
\ref{sec:racnn} describes our radius-adaptive convolution scheme; section
\ref{sec:results} presents the results, and section \ref{sec:conclusion}
concludes the paper.

\section{Related Work}
\label{sec:related_work}

The works that trying to reduce the convolution overhead can be
roughly divided into two categories: weight thinning and network thinning.
In the weight thinning the idea is to decrease the number of weights or bits-per-weights. 
Many schemes have been introduced that fit into this category.
In the weight quantization technique, the idea is to utilize more computing
resources by quantizing the weights into low-bit numbers.
In \cite{vanhoucke2011improving} 8-bit parameters have been used 
and results show a speedup with small loss of accuracy.
Authors in \cite{gupta2015deep} used 16-bit fixed-point which
results in significant memory usage and floating-point operation reduction with
comparable classification accuracy.
In the extreme case of the 1-bit per weight or binary weight neural
networks, the main idea is to directly train binary weights
\cite{courbariaux2015binaryconnect,courbariaux2016binarized,rastegari2016xnor}.

The more commonly used technique in weight thinning is to
decrease the parameters or neurons for each layer. In addition to network
complexity, it also can address the over-fitting. 
One effective approach to reducing weights is analyzing pre-trained
weights in a CNN model and remove non-informative ones.
For example, \cite{han2015learning} proposed a method to reduce the total number
of parameters and operations for the entire network. 
Authors in \cite{srinivas2015data} searched for the redundancy among neurons
and proposed a data-free pruning method to remove redundant neurons. 
HashedNets model proposed in \cite{chen2015compressing}
used a low-cost hash function to group weights into hash
buckets for parameter sharing. 
\cite{wen2016learning} utilized a structured sparsity regularizer in each layer
to reduce trivial filters, channels or even layers.
\cite{ullrich2017soft} proposed a regularization method based
on soft weight-sharing, which included both quantization and pruning in one
training procedure.
\cite{howard2017mobilenets} introduces an efficient family of CNNs called
MobileNet, which uses depth-wise separable convolution operations to drastically
reduce the number of computations required and the model
size. Other extensions to this work have tried to improve the speed and accuracy
\cite{sandlermobilenetv2,howard2019searching}.
Low-rank factorization technique uses matrix decomposition to
estimate the informative parameters of the deep CNNs. Therefore, non-informative weights 
can be removed to reduce the parameter to save computation.
In \cite{tai2015convolutional} authors proposed a new method for training
low-rank constrained CNNs based on the decomposition.
They used batch normalization is used to transform the activation of the
internal hidden units.
Using the dictionary learning
idea, learning separable filters was introduced
by \cite{rigamonti2013learning}. Results in \cite{denton2014exploiting} 
show considerable speedup for a single convolutional layer. They used low-rank
approximation and clustering schemes for the convolutional
kernels. The authors in \cite{jaderberg2014speeding} aim for a speedup in text
recognition. They used tensor decomposition schemes and their results show a
small drop in the accuracy.

In the network thinning, the idea is to compress the deep and wide
networks into shallower ones. Recently methods have adopted
knowledge distillation to reach this goal \cite{ba2014deep}.
\cite{hinton2015distilling} introduced a knowledge distillation
compression framework, which simplifies the training of deep networks by
following a student-teacher paradigm. Despite its simplicity, it achieves
promising results in image classification tasks. 
There are other approaches that utilize other speedup techniques such as FFT
based convolutions \cite{mathieu2013fast} or fast convolution using the Winograd
algorithm \cite{lavin2016fast}.
Stochastic spatial sampling pooling also is another speedup
idea used in \cite{zhai2017s3pool} based on the idea of inverse bilateral
filters \cite{saeedan2018detail}. Because of their complexity in the
implementation, these approaches also are not as commonly used in practical
solutions except for specific applications.

In both weight and network thinning the total number of weights is reduced,
while in the proposed RACNN, conversely, the main focus is avoiding any decrease
in the number of weights. Another advantage of RACNN is the simplicity of the
implementation. It can be implemented easily as a form of two convolutions. 

Recently, content-adaptive has become an active research topic. 
and they have been used 
in several task-specific use cases, such as and Monte
Carlo rendering denoising \cite{bako2017kernel}, motion prediction
\cite{xue2016visual} and semantic segmentation \cite{harley2017segmentation}.
Dynamic filter networks is an example of content-adaptive filtering techniques
\cite{jia2016dynamic}.
Filter weights are themselves directly predicted by a separate network branch,
and provide custom filters specific to different input data.
An extension of this work in \cite{wu2018dynamic} also learns from
multiple neighboring regions using position-specific kernels.
Authors in \cite{dai2017deformable} propose
deformable convolution, which produces position-specific
modifications to the filters.
Pixel-adaptive convolution proposed by \cite{su2019pixel} is another example of
content-adaptive convolution. In this work, filter weights are multiplied with
a spatially varying kernel that depends on learnable, local pixel features. 
In these adaptive approaches usually, the main goal is not addressing the
complexity which is the case for RACNN.

\section{Proposed Framework}
\label{sec:racnn}

In this section, we first describe the general criteria in
our speedup method. Then, we explain our radius-adaptive convolution 
structure and finally, we discuss other similar adaptive ideas.

\subsection{General criteria}
\label{subsec:constraints}

In this work, we were seeking for a speed optimization solution that meets four
criteria. First, the method, regardless of the speed, should be able to be
implemented by standard deep learning libraries such as \textit{TensorFlow},
\textit{PyTorch}, and \textit{Keras}. Therefore, the unoptimized code can be run 
on a standard library. This facilitates the training and testing which will
not require any low-level programming.
Second, the speed-optimized method should be able to be implemented by a general
matrix multiplication (GeMM). Deep learning libraries mostly use GeMMs, since,
they are fast and speed-wise optimized.
Next, the number of trainable parameters should not be decreased. Our goal
is to keep the number of neurons at the same level so the accuracy does not get
affected.
Finally, memory access also should be considered as a speedup factor in the
solution. Otherwise, for a typical computer, the speedup may become
insignificant.

\subsection{Radius-Adaptive Convolution}
\label{subsec:racnn}

\begin{figure}[t]
\begin{center}
   \includegraphics[width=0.95\linewidth]{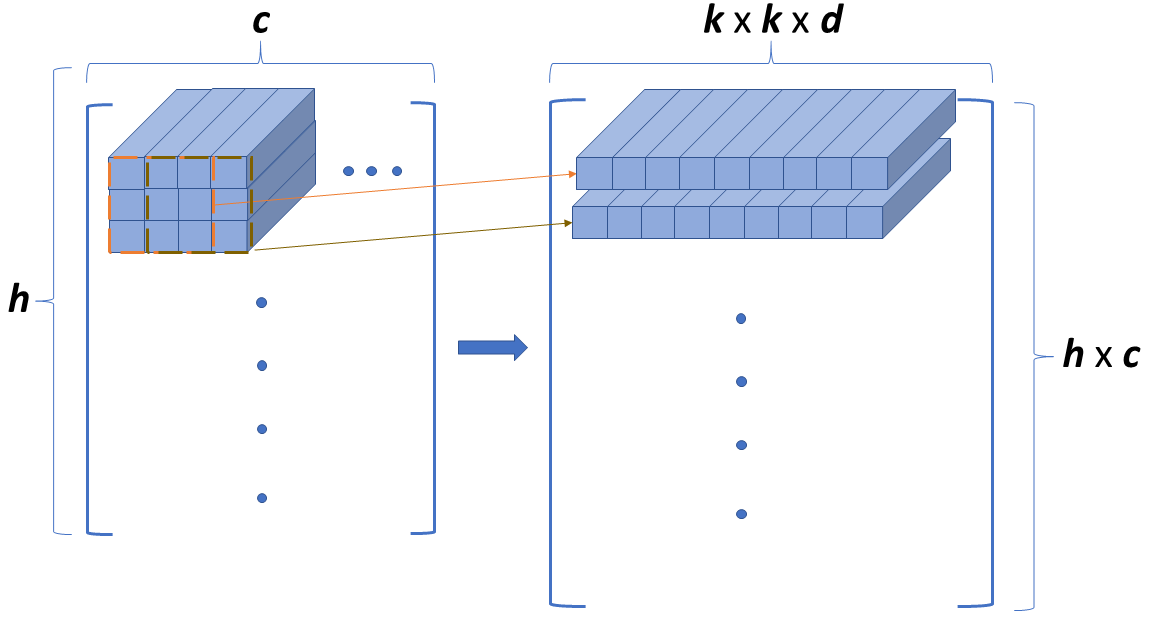}
\end{center}
   \caption{Visual illustration of an Image-to-column process that reshapes a
   3D input matrix into 2D when the kernel size $\boldsymbol{k}=3$.}
\label{fig:im2col}
\end{figure}

Convolutional layer usually is the most time-consuming layer of the network. Therefore, our idea
is to optimize convolutional layer algorithmically. With considering the
criteria in \ref{subsec:constraints} it seems a content-adaptive solution is reasonable. 
In a convolutional layer, a set of filters $\boldsymbol{w}_k$ is convolved by
the input. Assuming $\boldsymbol{k}$ is the kernel size,
$\boldsymbol{k}=2\boldsymbol{r}+1$, where $\boldsymbol{r}$ is the radius of the
kernel.  The input is in the form of a 3D matrix with the rows, columns, and
depth of $\boldsymbol{h}$, $\boldsymbol{c}$, and $\boldsymbol{d}$.
This input has $\boldsymbol{h}{\times}\boldsymbol{c}$ pixels and each
pixel has $\boldsymbol{d}$ channels. Assuming an image-to-column process can
reshape the 3D input data into a 2D form $\boldsymbol{I}_k$, so each row of
the matrix contains $\boldsymbol{k}{\times}\boldsymbol{k}$ pixels and there are
$\boldsymbol{h}{\times}\boldsymbol{c}$ rows (see Figure~\ref{fig:im2col}).
Therefore the convolution can be computed as
\begin{equation}
\label{eq:conv}
\boldsymbol{O} = \boldsymbol{I}_k \times \boldsymbol{w}_k\;, 
\end{equation}
where $\boldsymbol{w}_k$ has
$\boldsymbol{k}{\times}\boldsymbol{k}{\times}\boldsymbol{d}$ rows and
$\boldsymbol{f}$ columns and $\boldsymbol{f}$ is both the number of filters and
the depth of the output $\boldsymbol{O}$. 
The kernel radius $\boldsymbol{r}$ and therefore, the kernel size
$\boldsymbol{k}$ is fixed for the all pixels in the convolution.
Our idea is to have an adaptive radius. For simplicity let's consider  the two
options $\boldsymbol{r}=0$ and $\boldsymbol{r}=1$. Based on the value of each
pixel, either $\boldsymbol{r}=0$ or $\boldsymbol{r}=1$ will be selected.
In that case, the output will be computed as
\begin{equation}
\label{eq:conv_sep}
\boldsymbol{O}_a[p]= 
\begin{cases}
    \boldsymbol{I}_3[p]\times \boldsymbol{w}_3, \hat{\boldsymbol{r}}[p]=1& \\
    \boldsymbol{I}_1[p]\times \boldsymbol{w}_1, \hat{\boldsymbol{r}}[p]=0& 
\end{cases} , 
\end{equation}
where $\boldsymbol{w}_3$ and $\boldsymbol{w}_1$ are the $3{\times}3$ and
$1{\times}1$ convolution kernels with $\boldsymbol{r}=1$ and $\boldsymbol{r}=0$.
Assuming $[{\cdot}]$ is an indexing operator, 
$[p]$ shows the row $p$ in the matrix $\boldsymbol{I}_k$ and
$\boldsymbol{O}_a$. $\hat{\boldsymbol{r}}$ has a value of either 0 or one
depending on the radius of the kernel.
In (\ref{eq:conv_sep}) for each row $p$, there is a matrix multiplication. If we
split the input into two $\hat{\boldsymbol{I}}_3$ and $\hat{\boldsymbol{I}}_1$
matrices based on the radius we can compute each convolution separately by two
matrix multiplication as
\begin{equation}
\label{eq:conv_sep2}
\begin{array}{ll}
& \boldsymbol{O}_3 = \hat{\boldsymbol{I}}_3 \times \boldsymbol{w}_3 \\
& \boldsymbol{O}_1 = \hat{\boldsymbol{I}}_1 \times \boldsymbol{w}_1 
\end{array}, 
\end{equation}
and the results can be achieved by merging two outputs as
\begin{equation}
\label{eq:conv_sep3}
\boldsymbol{O}_a[p]= 
\begin{cases}
    \boldsymbol{O}_3 \left [ \boldsymbol{M} \left [ p \right ] \right ],
    \hat{\boldsymbol{r}}[p]=1&
    \\
    \boldsymbol{O}_1 \left [ \boldsymbol{M} \left [ p \right ] \right ],
    \hat{\boldsymbol{r}}[p]=0& \end{cases} , 
\end{equation}
where $\boldsymbol{M}$ is an index mapping table that maps
pixels in $\boldsymbol{O}_3$ and $\boldsymbol{O}_1$ to $\boldsymbol{O}_a$.
Algorithm \ref{alg:split} shows how to split the input
into $\hat{\boldsymbol{I}}_3$ and $\hat{\boldsymbol{I}}_1$ and obtain
$\boldsymbol{M}$ at the same time.
\begin{algorithm}
\small
\caption{Splitting input based on radius}
\label{alg:split}
\begin{algorithmic}[1]
\State \textbf{Results:} $\boldsymbol{M}$, $\hat{\boldsymbol{I}}_3$,
$\hat{\boldsymbol{I}}_1$ \State $i_1\gets 0$, $i_3\gets 0$, $p\gets 0$
\While{$p<h{\times}c$}
\If{$\hat{\boldsymbol{r}}[p]$=1}
\State $\hat{\boldsymbol{I}}_3[i_3]\gets \boldsymbol{I}_3[p]$
\State $\boldsymbol{M}[p]\gets i_3$
\State $i_3\gets i_3+1$
\Else
  \State $\hat{\boldsymbol{I}}_1[i_1]\gets \boldsymbol{I}_1[p]$
\State $\boldsymbol{M}[p]\gets i_1$
\State $i_1\gets i_1+1$
\EndIf
\State $p\gets p+1$
\EndWhile
\end{algorithmic}
\end{algorithm}
 
\begin{figure}[t]
\begin{center}
   \includegraphics[width=0.5\linewidth]{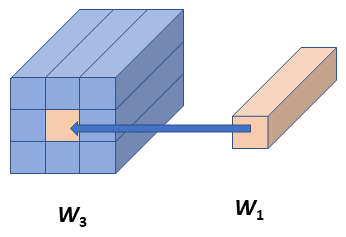}
\end{center}
   \caption{A visual demonstration of sharing weights between $3{\times}3$ and
   $1{\times}1$ convolution kernels.}
\label{fig:share}
\end{figure}

For the pixels with $\hat{\boldsymbol{r}}[p]=0$, the idea in
(\ref{eq:conv_sep2}) and (\ref{eq:conv_sep3}) result in a considerable reduction
in both operation and memory access, since both $\hat{\boldsymbol{I}}_1$
and $\boldsymbol{w}_1$ are $3{\times}3$ smaller than $\boldsymbol{I}_3$ and
$\boldsymbol{w}_3$. Consequently, a significant speedup can be achieved.
However, its disadvantage is the hard decision based on
$\hat{\boldsymbol{r}}[p]$ value.
This, causes a discontinuity in the output and makes it untrainable for standard
gradient-based optimization algorithms. To solve this problem we propose a soft
decision, based on a linear combination of both matrices. 
This modifies (\ref{eq:conv_sep3}) to
\begin{equation}
\label{eq:conv_sep4}
\boldsymbol{O}_a[p]= 
    \boldsymbol{O}'_3[p]\cdot \alpha[p] +
    \boldsymbol{O}'_1[p]\cdot(1-\alpha[p]).
\end{equation}
$\boldsymbol{O}'_3=\boldsymbol{I}_3{\times}\boldsymbol{w}_3$ and
$\boldsymbol{O}'_1=\boldsymbol{I}_1{\times}\boldsymbol{w}_1$ are the
$3{\times}3$ or $1{\times}1$ convolutions. 
For each pixel (or row) in the $\boldsymbol{I}_k$,
$0{\leq\alpha[p]\leq}1$ determines how much of each convolution contributes to the output
$\boldsymbol{O}_a$. In the case of $\alpha[p]=1$ and $\alpha[p]=0$ either of
$3{\times}3$ or $1{\times}1$ will be selected. The problem, however,
is when $0<\alpha[p]<1$. This means both $3{\times}3$ and $1{\times}1$ should be
computed for that pixel. This not only contradicts the speedup idea but also
adds some redundancy since two convolutions with different radii will be
computed for the same pixel. To solve this problem we propose to make weights
at the center of $\boldsymbol{w}_3$ equal to $\boldsymbol{w}_1$ (see
Figure~\ref{fig:share}).
Let's consider $\boldsymbol{w}_3$ as a 4D convolution
kernel with height, width, depth, and filter number equal to 3, 3,
$\boldsymbol{d}$ and, $\boldsymbol{f}$ respectively, thus,
\begin{equation}
\label{eq:conv_share}
    \boldsymbol{w}_3[1,1,:,:]=\boldsymbol{w}_1 .
\end{equation}
$\boldsymbol{w}_3[1,1,:,:]$ points to $\boldsymbol{w}_3$ at the row and column
of 1 and 1 which has a dimension of
$1{\times}1{\times}\boldsymbol{d}{\times}\boldsymbol{f}$.
By sharing the weights between two convolutions we can address the redundancy problem. If we
substitute $\boldsymbol{O}_3=\boldsymbol{I}_3{\times}\boldsymbol{w}_3$ and
$\boldsymbol{O}_1=\boldsymbol{I}_1{\times}\boldsymbol{w}_1$ in
(\ref{eq:conv_sep4}) we get
\begin{equation}
\label{eq:conv_final}
\boldsymbol{O}_a= 
    \boldsymbol{I}_1{\times}\boldsymbol{w}_1 + \alpha \cdot
    \boldsymbol{I}_3{\times}(\boldsymbol{w}_3-\boldsymbol{w}_1) \;,
\end{equation}
where, $(\boldsymbol{w}_3-\boldsymbol{w}_1)$ is a hollow $3{\times}3$
kernel.
In other words, a kernel that its weights at the center are zero.
Note that, $\boldsymbol{w}_1$ should be padded with zeros to have the same size
as $\boldsymbol{w}_3$.
In (\ref{eq:conv_final}), the number of operations and memory access is in
the worst-case scenario is equal to a normal convolution. Worst case scenario
happens when the $\alpha{>}0$ for all pixels in the input so the
$\boldsymbol{I}{\times}(\boldsymbol{w}_3-\boldsymbol{w}_1)$ convolution should
be computed for all pixels. Otherwise, we save some computation time by skipping
$3{\times}3$ convolution for some of the pixels.
In (\ref{eq:conv_final}), for each pixel $\alpha$ should be calculated
depending on the content of the pixel. Consequently, the network should learn
and calculate the $\alpha$. One way to this is utilizing a $1{\times}1$ convolution.
$1{\times}1$ kernel $\boldsymbol{w}_{\alpha}$ is convolved by the input
$\boldsymbol{I}_1$ as
\begin{equation}
\label{eq:alpha}
\alpha= 
    \textbf{Min} \left (
    \textbf{Max} \left ( \boldsymbol{I}_1{\times}\boldsymbol{w}_{\alpha},0
    \right ),1 \right ).
\end{equation}
The output of convolution should be clipped to ensure values are between zero
and one. Since, the output has only one value, the number of filters
$\boldsymbol{f}$ in $\boldsymbol{w}_{\alpha}$ is one. For maximum speedup, our
goal is to have the minimum number of matrix multiplication. By observing
(\ref{eq:alpha}) and the first term in (\ref{eq:conv_final}), we realize we can
merge two matrix multiplication into one since both have the same input as
\begin{equation}
\label{eq:alpha_weight}
(\alpha'|\boldsymbol{O}'_1)= \boldsymbol{I}_1{\times} (
\boldsymbol{w}_{\alpha}|\boldsymbol{w}_1)\:, 
\end{equation}
where $\alpha'=\boldsymbol{I}_1{\times}\boldsymbol{w}_{\alpha}$. 
Once we have $\alpha$, which is the clipped $\alpha'$, we find
the output $\boldsymbol{O}_a$ as
\begin{equation}
\label{eq:output}
\begin{split}
\boldsymbol{O}_a & = 
     \boldsymbol{O}'_1 + \alpha \cdot 
     \boldsymbol{I}'_3{\times}\boldsymbol{w}'_3 \\
    \boldsymbol{I}'_3 & = \{\boldsymbol{I}_3[p]\:|\:\alpha[p]{>}0\}
\end{split}\:.
\end{equation}
$\boldsymbol{I}'_3$ is a subset of $\boldsymbol{I}_3$ only for the rows with
$\alpha[p]{>}0$ and $\boldsymbol{w}'_3=\boldsymbol{w}_3{-}\boldsymbol{w}_1$ is
a hollow kernel (see Figure~\ref{fig:main}).
The speedup factor depends on the number of pixels with $\alpha{=}0$.
Let's consider an example, where $\alpha{=}0$ for 50\% of the pixels.
Theoretically, the computation time of (\ref{eq:alpha_weight}) and
(\ref{eq:output}) are $\frac{1}{9}$ and $\frac{8}{9}$ of a standard
$3{\times}3$ convolution. In this example time becomes
$\frac{1}{9}+\frac{1}{2}{\times}\frac{8}{9}=0.55$ of a standard convolution.
When $\alpha{=}1$ for all pixels, radius-adaptive convolution is equivalent
to a standard convolution.

\subsection{Other Adaptive Ideas}
\label{subsec:adaptive}
 
 we also analyzed similar content-adaptive ideas such as a convolution with
 an adaptive number of filters. Similar to RACNN, we can split the input into
 two matrices and each gets multiplied by different weights with different
 filter sizes. One major problem is the hard-decision, however, prior to
 solving that, we realized even in the best-case scenario we cannot get a
 satisfactory speedup.
 The main reason is, increasing or decreasing the size of filters
 $\boldsymbol{f}$ in $\boldsymbol{w}$ does not affect the speed in the
 convolution significantly. Due to the number of memory access, 
 the bottleneck is usually the big input matrix $\boldsymbol{I}$.
 Although, this feature is a drawback in this idea but, it's an advantage in the
 RACNN, because it makes the overhead cost of computing the $\alpha$ in
 (\ref{eq:alpha_weight}) minor.

\section{Experimental Results}
\label{sec:results}
We considered an object classification network to test the accuracy and
speed of RACNN. We selected two well-known image recognition graphs VGG16
\cite{simonyan2014very} and ResNet50 \cite{he2016deep} as a base to analyze the
accuracy and speed.
For these tests we used COCO-2017 \cite{lin2014microsoft} training and
validation dataset with approximately 850000 and 36000 object images and 80 classes.
In the following, we first present the accuracy results then we discuss
the speed.

\subsection{Accuracy}
\label{sec:accuracy}

\begin{figure}
\begin{center}
   \includegraphics[width=0.98\linewidth]{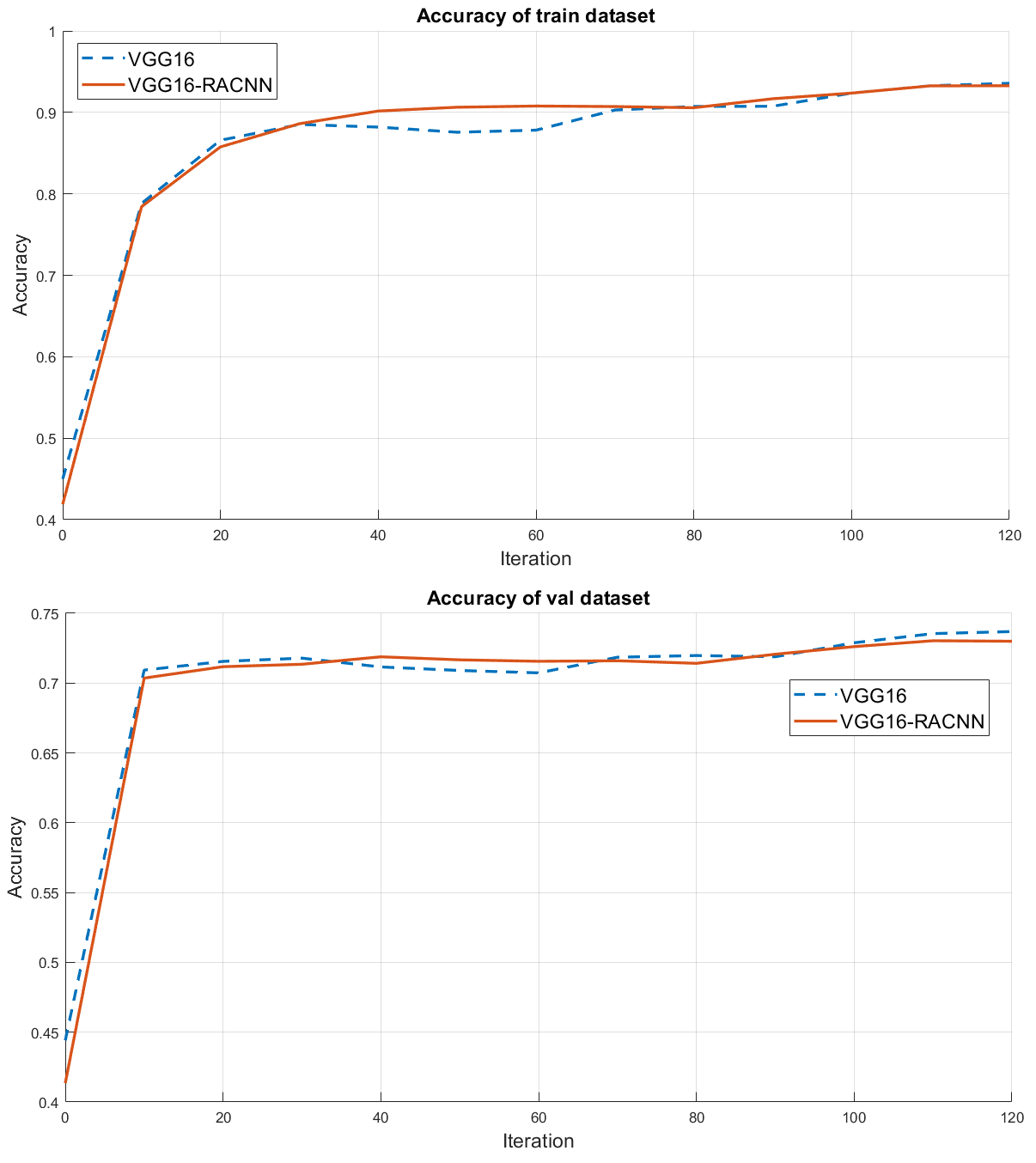}
\end{center}
   \caption{Accuracy comparison for VGG16 graph with and without RACNN. Top is
   training and bottom is the validation dataset.
   We expect similar accuracies for both.}
\label{fig:vgg_acc}
\end{figure}

\begin{figure}
\begin{center}
   \includegraphics[width=0.98\linewidth]{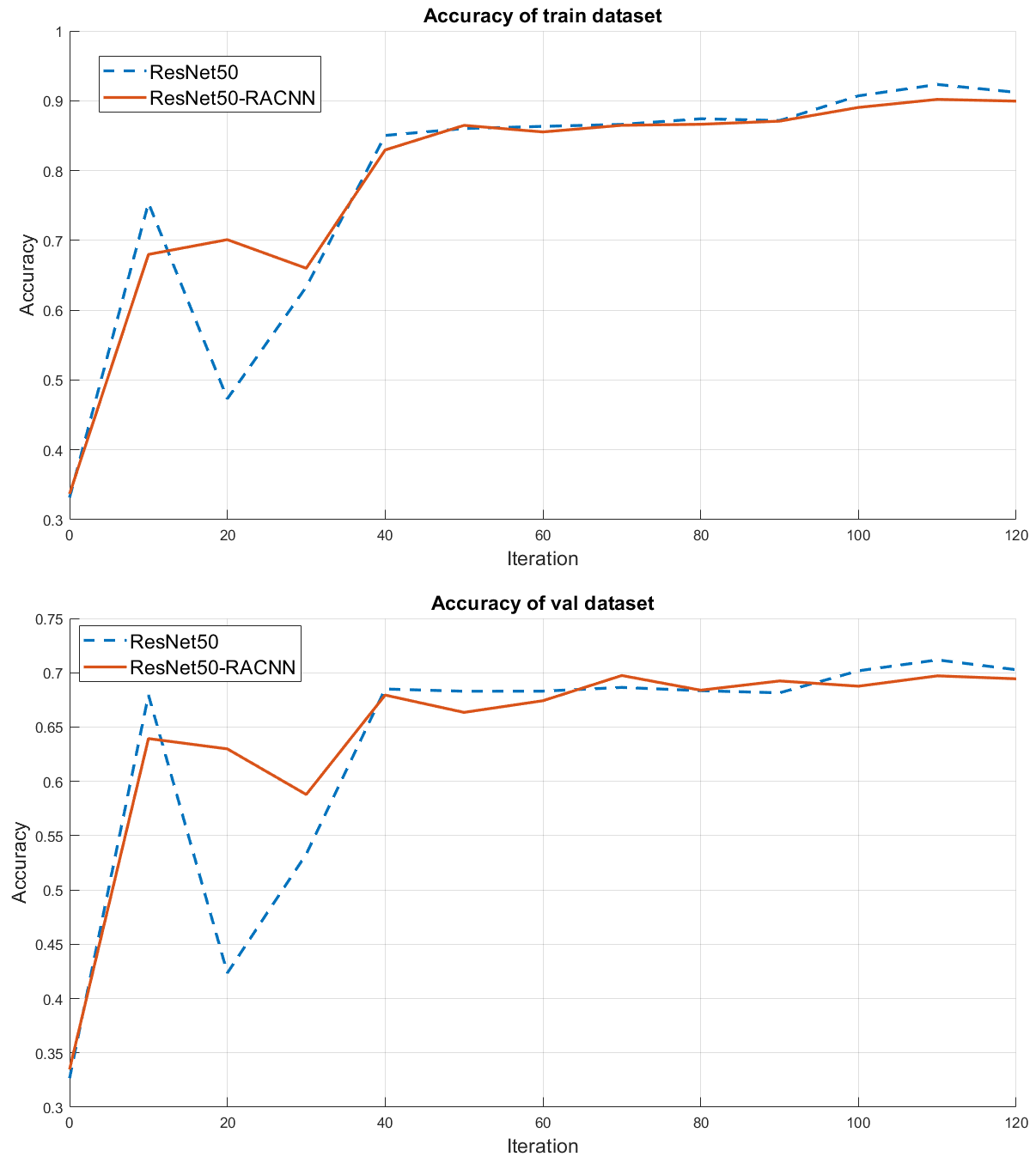}
\end{center}
   \caption{Accuracy comparison for ResNet50 graph with and without RACNN. Top
   is training and bottom is the validation dataset.
   We expect similar accuracies for both.}
\label{fig:resnet_acc}
\end{figure}

To measure the efficiency of RACNN in accuracy, our idea is to replace the
$3{\times}3$ convolutions with radius-adaptive ones and examine the results.
These modifications led to VGG16-RACNN and ResNet50-RACNN new graphs.
For VGG16, the computation time of deep layers is negligible compared to the
first layers. This is due to a small resolution of images at deep layers.
Therefore, accelerating those layers won't affect the total computation
time of the network. Thus, for VGG16-RACNN, we set the first 7 convolutions
(or the first 3 stages) as radius-adaptive convolutions and we kept others as
standard convolutions.
For ResNet, however, we replaced all 16 convolutions.
For ResNet, even though the resolution in deep layers is low, the number of
filters is relatively high.
Then, we used COCO-2017 training dataset to train all 4 graphs (i.e., VGG16,
ResNet50, VGG16-RACNN and ResNet50-RACNN). We used Adam optimizer with a
learning rate of 1e-4 and 120 iterations. The batch size was set to maximum possible for
our hardware which was 30 for VGG16 and VGG16-RACNN and 40 for ResNet50 and
ResNet50-RACNN. 
After each iteration, we calculated the classification accuracy for both graphs
using both training and validation datasets. The advantage of RACNN is that the
unoptimized code can be implemented easily by standard libraries. For training
and accuracy calculation we used Keras library in Python.
Figures \ref{fig:vgg_acc} and \ref{fig:resnet_acc} compare the accuracy of both
graphs with and without RACNN. We expect to get similar accuracies since both
networks (i.e., standard and adaptive) have similar architecture and number of
parameters.
Besides in case of $\alpha{=}1$ radius-adaptive convolution is equivalent to a
standard convolution (see Figure~\ref{fig:main}). The results also confirm this fact.
Although they may not follow a same learning path, they converge to same
accuracy.

\begin{figure}
\begin{center}
   \includegraphics[width=0.98\linewidth]{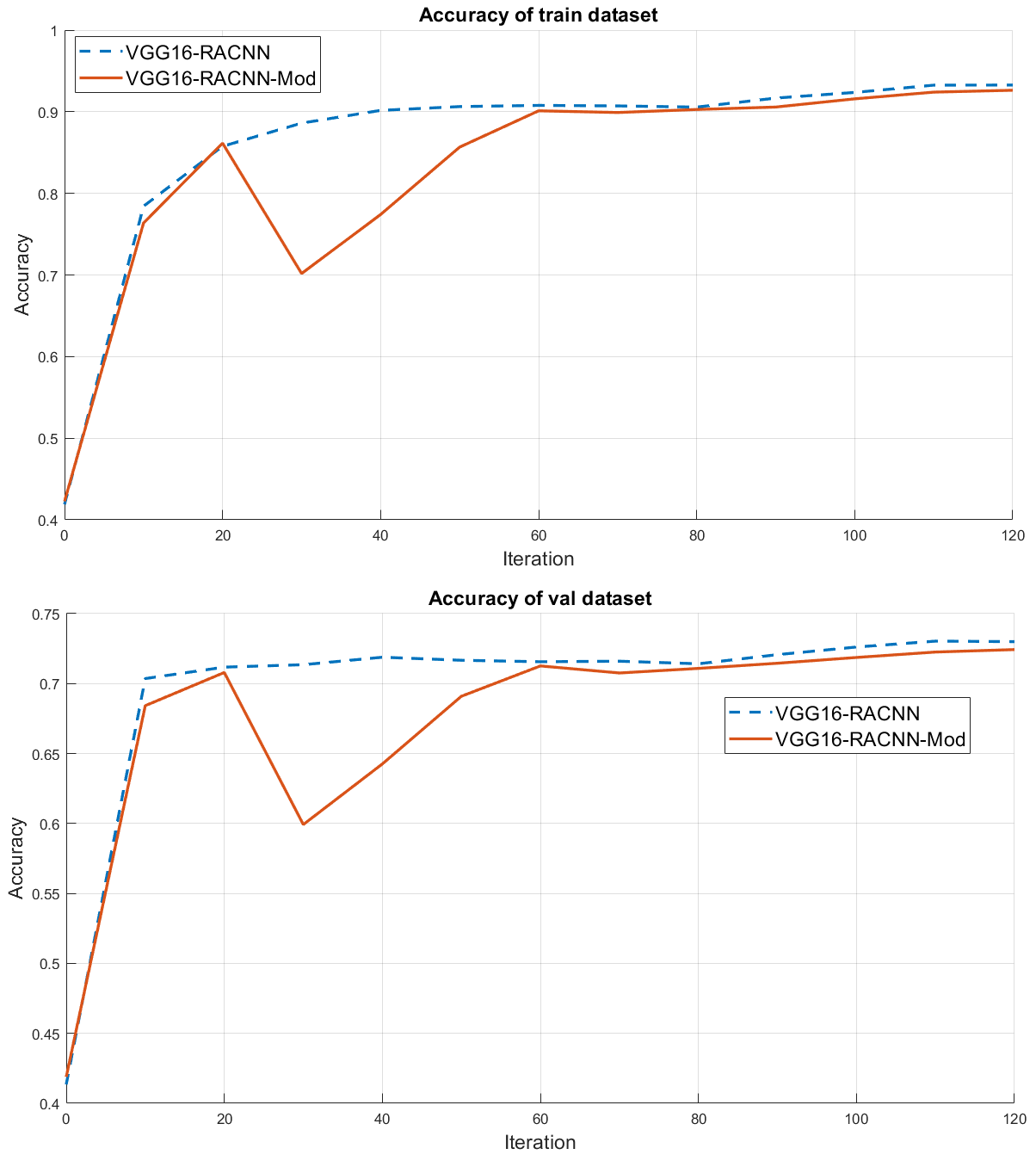}
\end{center}
   \caption{Accuracy comparison by replacing one layer of VGG16-RACNN with
   a standard convolution for training (top) validation (bottom)
   datasets.
   We expect similar accuracies for both.}
\label{fig:vgg_acc_mod}
\end{figure}

Another interesting test is examining the effect of RACNN for only one
convolution in the network. For this test, we replaced only the first
adaptive layer of VGG16-RACNN with a standard convolution layer and measured the
accuracy. Figure~\ref{fig:vgg_acc_mod} compares the results for 120 iterations
which also confirms a similarity in the accuracy.

\subsection{Speed}
\label{sec:speed}
\begin{figure}
\begin{center}
   \includegraphics[width=0.98\linewidth]{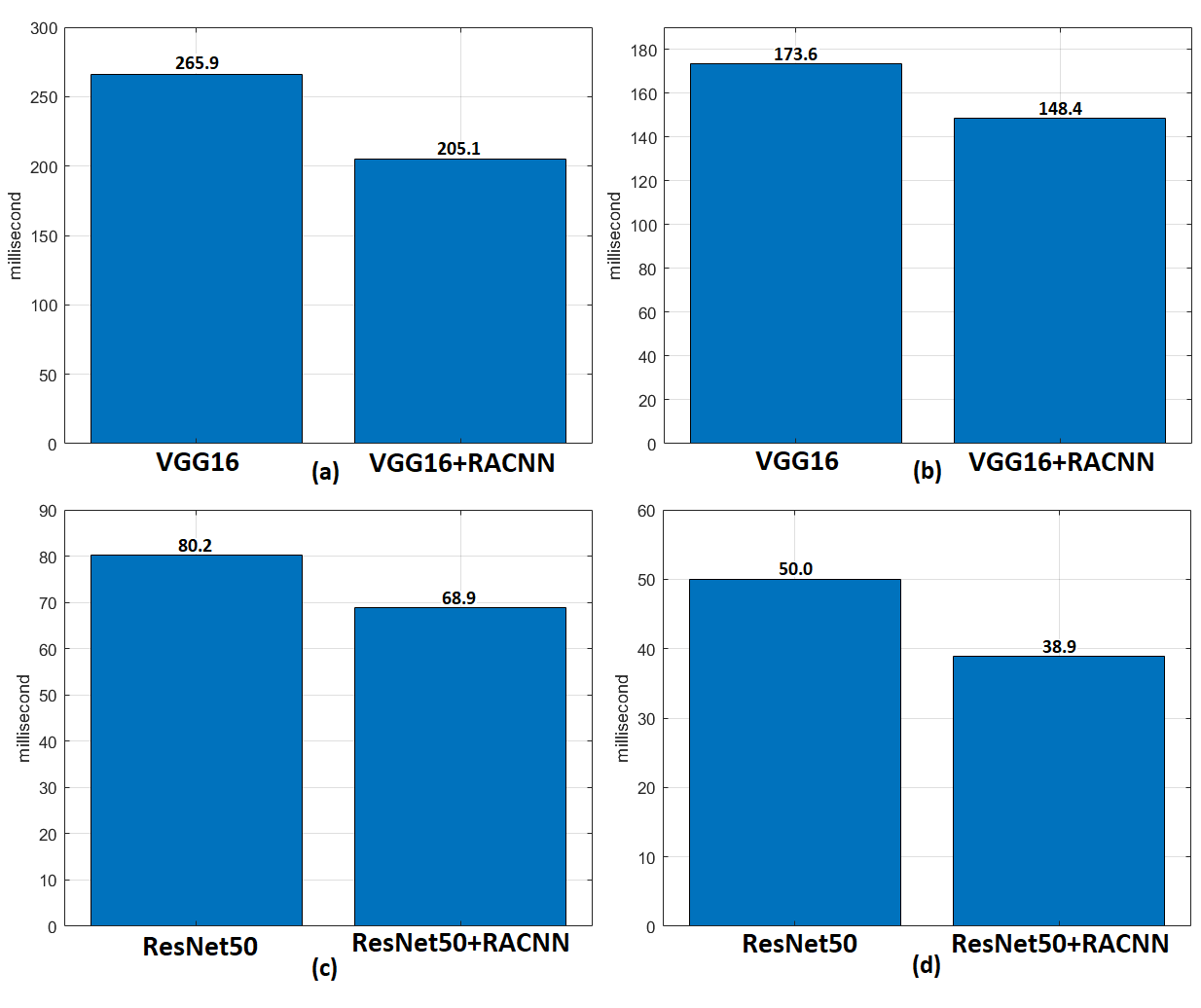}
\end{center}
   \caption{Average of computation time in milliseconds for 1000 samples for two
   graphs VGG16 (top) and Resnet50 (bottom) using two CPUs i7-6700 (left) and 
   i7-8750 (right).}
\label{fig:speed}
\end{figure}

Now that we confirmed the similarity of accuracies between the proposed and
standard models, we need to compare the speed results. Unlike accuracy, the
speed cannot be tested with high-level programming using standard libraries.
In order to optimally implement our custom design RACNN, we needed to use a
lower level of programming. To have a fair comparison we had to also implement
the standard graphs in the same way.
We used Python to implement our code. For matrix multiplication, and other
general operations, we used Numpy and cuBLAS for CPU and GPU.
For other customized tasks, such as splitting and merging in
(\ref{eq:output}), we used C++ and CUDA for CPU and GPU and we designed
high-level interfaces for them to be used in Python.
Then, we measured the computation time for all graphs using two different CPUs;
i7-6700@2.6GHz and i7-8750H@2.2GHz.
We ran the object classification network for the first 1000 object images in the
COCO-2017 validation dataset and we computed the average. Figure \ref{fig:speed}
compares the average computation time between the graphs with and without RACNN.
Figure \ref{fig:speed} shows a 23\% speed improvemnt on average for all tests.

We also tested the speed with GPU using Nvidia GTX 1050. We an observed
insignificant speedup (4\%). We believe, with more optimized code for
kernels that handle splitting and merging in (\ref{eq:output}) we can reach
speeds closer to the CPU results.

\begin{figure}
\begin{center}
   \includegraphics[width=0.98\linewidth]{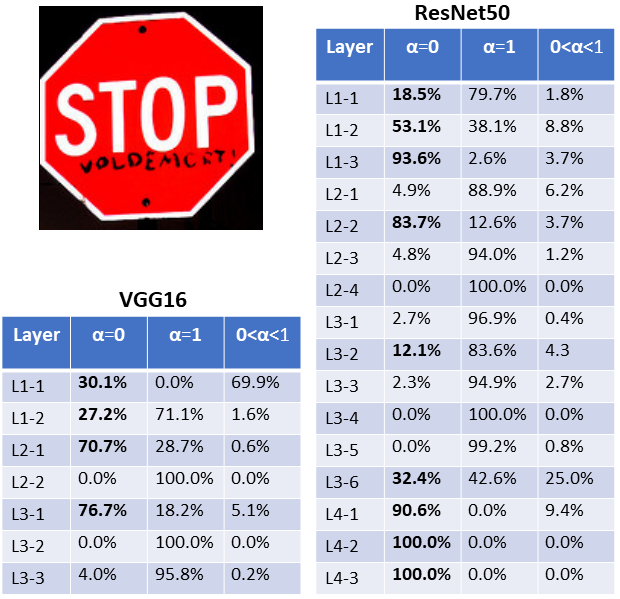}
\end{center}
   \caption{Percentage of $\alpha$ for different values for a sample image.
   Higher percentage in $\alpha=0$ leads to higher speedup.}
\label{fig:t1}
\end{figure}

\begin{figure}
\begin{center}
   \includegraphics[width=0.98\linewidth]{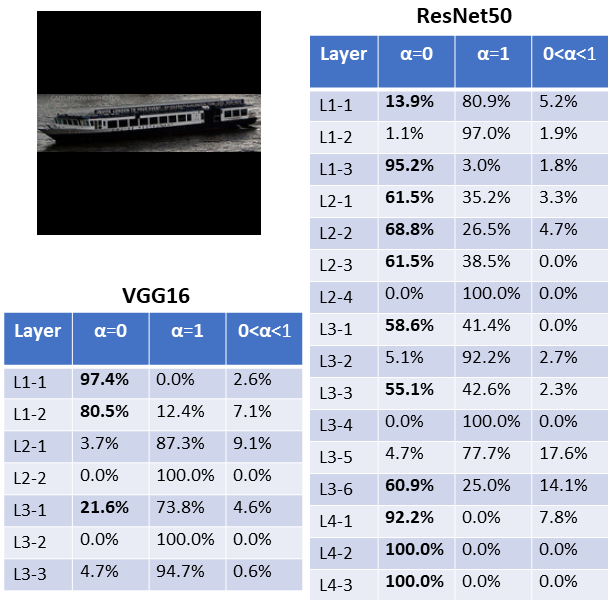}
\end{center}
   \caption{Percentage of $\alpha$ for different values for a sample image.
   Higher percentage in $\alpha=0$ leads to higher speedup.}
\label{fig:t2}
\end{figure}

To more deeply analyze the effectiveness of RACNN, we measured the contribution
of it in each convolution layer for two sample images. The percentage of the
input pixels with $\alpha=0$, defines the amount of speed-up by RACNN.
Figures~\ref{fig:t1} and \ref{fig:t2} show the percentage of $\alpha$
for sample images and we highlighted the more than 10\%
which includes the majority of the layers. By analyzing these numbers, we can
possibly adjust the RACNN graph. If there a consistent low-percentage
contribution for a specific layer, that layer we can be replaced with a standard
convolution to avoid the overhead cost of $\alpha$ calculation.
   
\section{Conclusion}
\label{sec:conclusion}

In this paper, we presented a content-adaptive convolution that can reduce the
number of operations and memory access in a convolution layer without decreasing the number of trainable parameters.   
Our proposed radius-adaptive
convolution utilizes different radii
based on the content. We implemented RACNN
and tested the results for both CPU and GPU. As we expected, the accuracy of
RACNN is similar to a standard convolution. Results also show a significant
speedup for CPU implementation. The speedup gain in GPU, however, is lower
than CPU, and more works need to be done to make the code as efficient as CPU.

{\small
\bibliographystyle{ieee_fullname}
\bibliography{egbib}
}

\end{document}